\documentclass[runningheads]{llncs}

\usepackage{llncsdoc}

\usepackage{amsmath}
\usepackage{latexsym}
\usepackage{amssymb}
\usepackage{mathtools}

\usepackage{epsf}
\usepackage[all]{xy}
\usepackage{verbatim}
\usepackage{times}
\usepackage{color}

\usepackage{tikz}
\usepackage{pgfplots}
\usetikzlibrary{trees}
\usetikzlibrary{topaths}
\usetikzlibrary{decorations.pathmorphing}
\usetikzlibrary{decorations.markings}
\usetikzlibrary{matrix,backgrounds,folding}
\usetikzlibrary{chains,scopes,positioning,fit}
\usetikzlibrary{arrows,shadows}
\usetikzlibrary{calc} 
\usetikzlibrary{chains}
\usetikzlibrary{shapes,shapes.geometric,shapes.misc}

\pgfdeclarelayer{edgelayer}
\pgfdeclarelayer{nodelayer}
\pgfsetlayers{background,edgelayer,nodelayer,main}

\tikzstyle{dot}=[circle,fill=black,draw=black]
\tikzstyle{every picture}=[baseline=(current bounding box).east,scale=0.5,node distance=5mm]
\tikzstyle{none}=[inner sep=0pt]
\tikzstyle{every loop}=[]

\tikzstyle{(null)}=[]
\tikzstyle{plain}=[]

\usepackage{latexsym,lingmacros}

\title{Static and Dynamic Vector Semantics for Lambda Calculus Models of Natural Language}
\author{Mehrnoosh Sadrzadeh and Reinhard Muskens}

\institute{
School of Electronic Engineering and Computer Science,\\ 
Queen Mary University of London.\\
\email{mehrnoosh.sadrzadeh@qmul.ac.uk}\\
Department of Philosophy, Tilburg University.\\
\email{ r.a.muskens@gmail.com }
}

\titlerunning{Context Update  for Lambdas and Vectors}

\authorrunning{M.\ Sadrzadeh and R.\ Muskens}

\newcommand{\ov}{}
\newcommand{\ovl}{\overline}

\begin{document}
\maketitle

\begin{abstract}
Vector models of language are based on the contextual aspects of language, the distributions of words 
and how they co-occur in text. Truth conditional models focus on the
logical aspects of language,  compositional properties of  words and how they compose to form  sentences. In the truth conditional approach,  the denotation of a
sentence determines its truth conditions, which  can be taken to be a
truth value, a set of possible worlds, a context change
potential, or similar. In the vector models,  the degree of co-occurrence  of words in context determines how similar the meanings of words are.  In this  paper, we put these two models together and develop a vector semantics for language based on the simply typed lambda calculus models of natural language. We provide two  types of vector  semantics: a static one that  uses techniques familiar from the truth conditional tradition
and a dynamic one  based on a form of dynamic interpretation inspired by Heim's context change potentials.  We show how the dynamic model can  be applied to   entailment between a corpus and a sentence and provide examples. 
\end{abstract}

\section{Introduction} 
Vector semantic models, otherwise known as distributional models, are
 based on the contextual aspects of language, the company each word
 keeps, and patterns of use in corpora of documents. Truth conditional
 models focus on the logical and denotational aspects of language, how
 words can be represented by functions over sets and application and
 composition of these functions. Vector semantics and truth
 conditional models are based on different philosophies; one takes the
 stance that language is contextual, the other asserts that it is
 logical.  In recent years there has been much effort to bring these
 two together. We have models that are based on a certain type of
 grammatical representation, e.g. the pregroup model of
 \cite{Coeckeetal}, the Lambek Calculus model of
 \cite{CoeckeGrefenSadr13}, and the combinatorial categorial models 
 of \cite{Krishnamurthy2013,Maillard2014}; we also have more concrete
 models that draw inspirations from type theory but whose major
 development is developing concrete ways of constructing linear and
 multi linear algebraic counterparts for the syntactic types, e.g.
 matrices and tensors of \cite{GrefenSadrCL,baroni-etal-2014} and
 relational clusters of \cite{Lewis-Steedman2013}.

What these approaches,
\cite{Coeckeetal,Krishnamurthy2013,Maillard2014} more than
\cite{baroni-etal-2014,Lewis-Steedman2013}, miss is acknowledging the
inherent gap between the contextual and truth conditional semantics;
they closely follow the truth theoretic conditions to assign vector
representations to phrases and sentences.  Indeed it is possible to
develop a stand-alone compositional vector semantics along these
lines, but this will result in a static semantics. From the
perspective of the underlying theory it will also be quite natural to
have such a vector semantics work in tandem with a dynamic theory and
let the two modules model different aspects of meaning. Distributional
semantics is particularly apt at modelling associative aspects of
meaning, while truth-conditional and dynamic forms of semantics are
good at modelling the relation of language to reality and at modelling
entailment. It is quite conceivable that a theory that combines the
two as separate modules will be simpler than one that tries to make
one of the approaches do things it was never intended for.

This is what we will do in this paper. We first sketch how an approach
to semantics that in many of its aspects derives from the one
pioneered by \cite{mont:prop74} can be used to assign vector meanings
to linguistic phrases. The theory will be based on the simply typed
lambda calculus and as a result will be neutral with respect to the
linguist's choice of syntax, in the sense that it can be combined with
any existing syntax-semantics interface that assumes that the
semantics is based on lambdas (e.g.\ linguistic trees + `type-driven
translation', f-structures + `glue' in Lexical-Functional Grammar, 
derivations in Combinatory Categorial Grammar + the use of combinators,
proofs + `semantic recipes' in Lambek Categorial Grammar, etc.).  

Our reasoning for the use of lambda calculus is that it directly
relates our semantics to higher order logic and makes standard ways of
treating long distance dependencies and coordination accessible to
vector-based semantics. This approach results in the same semantics as
the static approaches listed above. The reason for providing it is to
show that a lambdas calculus model of language can directly be
provided with a straightforward vector semantics. As will be seen,
abstract lambda terms, which can be used as translations of linguistic
expressions have a lot in common with the Logical Forms of these
expressions and the lambda binders in them give easy ways for treating
long distance dependencies. The use of lambda terms also makes
standard ways of dealing with coordination accessible to
distributional semantics. We provide extensive discussion of this and
will give examples where the direct use of lambdas gives an edge over
the above listed static approaches.

The above semantics does not have an explicit notion of context
however. The second contribution of this paper paper is that, based on
the same lambda calculus model of natural language, we develop a
dynamic vector interpretation for this type theory where denotations
of sentences are ``context change potentials", as introduced by Heim
\cite{Heim1983}. We show how to assign such a vector interpretation to
words and how these interpretations compose in a way such that the
vectors of the sentences containing them change the context in a
dynamic style similar to the one instructed by Heim. A context can be
interpreted in different ways, we work with two different notions of
context in distributional semantics: co-occurrence matrices and entity
relation graphs, encoded here in the form of cubes. Both of these are
built from corpora of documents and record the co-occurrence between
the words: in a simple neighbourhood window for the case of
co-occurrence matrices and in a window structured by grammatical
dependencies, in the case of an entity relation cube. We believe our
model is flexible enough with other distributional notions of
contexts, such as networks of grammatical dependancies. We show how
our approach relates to Heim's original notion of `files' as contexts.
Other dynamic approaches, such the update semantics of
\cite{Veltman1996} and the continuation-based semantics of
\cite{Groote2006}, can also be used; we aim to do this in the future.

We are after a compositional vector semantics, but this paper is
theoretical. So what we will not do is settle upon---from the armchair
so to speak---certain concrete representations of contexts and 
updates and a set of concrete vector composition operations for
combining phrases or concrete matrices or cubes that embody them. We
thus will leave an exhaustive empirical evaluation of our model to
future work, but show, by means of examples, how the notion of 
"admittance of sentences by contexts" from the context update logic of
Heim and Karttunen can be applied to develop a relationship between
matrix and cube contexts and sentences and how this notion can be
extended from a usual boolean relation to one which has degrees, based
on the notion of degrees of similarity between the words. This notion
resembles that of a "contextual entailment" between corpora and
sentences, we review the current entailment datasets that are
mainstream in distributional semantics and discuss how they can or
cannot be applicable to test this notion. but leave an experimental
valuation to future work. 

The lambda calculus approach we use is based on the Lambda Grammars of
\cite{cglfg,lll}, which were independently introduced as Abstract
Categorial Grammars (ACGs) in \cite{groote:acg}.  The theory developed
here, however, can be based on any syntax-semantics interface that
works with a lambda calculus based semantics.  Our approach is 
agnostic as to the choice of a syntactic theory.  Lambda Grammars/ACGs
are just a framework for thinking about type and term homomorphisms
and we are using them entirely in semantics here. 


This paper is the journal version of our previous extended abstract
\cite{Muskens-Sadr2016-1} and short paper \cite{Muskens-Sadr2016-2}.


\section{Lambda Grammars} 
The Lambda Grammars of \cite{cglfg,lll} were independently introduced
as Abstract Categorial Grammars (ACGs) in \cite{groote:acg}. An ACG
generates two languages, an \emph{abstract} language and an
\emph{object} language. The abstract language will simply consist of
all linear lambda terms (each lambda binder binds exactly one variable
occurrence) over a given vocabulary typed with \emph{abstract types}.
The object language has its own vocabulary and its own types. We give
some basic definitions here, assuming familiarity with the simply
typed $\lambda$-calculus.

If $\mathcal{B}$ is some set of basic types, we write
$\mbox{\emph{TYP}}(\mathcal{B})$ for the smallest set containing
$\mathcal{B}$ such that
$(\alpha\beta)\in\mbox{\emph{TYP}}(\mathcal{B})$ whenever
$\alpha,\beta\in\mbox{\emph{TYP}}(\mathcal{B})$. Let $\mathcal{B}_1$
and $\mathcal{B}_2$ be sets of basic types. A function $\eta$ from
$\mbox{\emph{TYP}}(\mathcal{B}_1)$ to
$\mbox{\emph{TYP}}(\mathcal{B}_2)$ is said to be a \emph{type
homomorphism} if $\eta(\alpha\beta)=(\eta(\alpha)\eta(\beta))$, for
all $\alpha,\beta\in\mbox{\emph{TYP}}(\mathcal{B}_1)$. It is clear
that a type homomorphism $\eta$ with domain
$\mbox{\emph{TYP}}(\mathcal{B})$ is completely determined by the
values of $\eta$ for types $\alpha\in\mathcal{B}$. 

Let us look at an example of a type homomorphism that can be used to
provide a language fragment with a classical Montague-like meaning.
Let $\mathcal{B}_1=\{D,N,S\}$ ($D$ stands for determiner phrases, $N$
for nominal phrases, $S$ for sentences), let $\mathcal{B}_2=\{e,s,t\}$
($e$ is for entities, $s$ for worlds, and $t$ for truth-values), and
let $h_0$ be defined by: $h_0(D)=e$,
$h_0(N)=est$,\footnote{Association in types is to the right and outer
parentheses are omitted; so $est$ is short for $(e(st))$, arguably a
good type for \emph{predicates}.} and $h_0(S)=st$. Then the types in
the second column of Table \ref{tab} have images under $h_0$ as given
in the fourth column.

\begin{table}[t]
\begin{center}
\def\arraystretch{1.4}\tabcolsep=5pt
\begin{tabular}{llll}\hline
constant $c$&type $\tau$&
$H_0(c)$&$h_0(\tau)$\\ \hline
\texttt{woman}&$N$& \textit{woman}&$est$\\
\texttt{man}&$N$& \textit{man}&$est$\\
\texttt{tall}&$NN$& \textit{tall}&$(est)est$\\
\texttt{smokes}&$DS$&\textit{smoke}&$est$\\
\texttt{loves}&$DDS$&\textit{love}&$eest$\\
\texttt{knows}&$SDS$&$\lambda p\lambda x\lambda w.
\forall w'(Kxww'\to pw')$&$(st)est$\\
\texttt{every}&$N(DS)S$&$\lambda P'\lambda P\lambda w.
\forall x(P'xw\to Pxw)$&$(est)(est)st$\\
\texttt{a}&$N(DS)S$&$\lambda P'\lambda P\lambda w.
\exists x(P'xw\land Pxw)$&$(est)(est)st$\\
\hline
\end{tabular}
\end{center}
\caption{\label{tab}An Abstract Categorial Grammar / Lambda Grammar 
connecting abstract terms with Montague-like meanings. Here $p$ is a
variable of type $st$, while $x$ is of type $e$, $w$ and $w'$ are of
type $s$, and $P$ and $P'$ are of type $est$. The constant $K$ of type
$ess$ denotes the epistemic accessibility relation.}
\end{table}
We now define the notion of a \emph{term homomorphism}. If $C$ is 
some set of typed constants, we write $\Lambda(C)$ for the set of all
lambda terms with constants only from $C$. The set of \emph{linear}
lambda terms over $C$ is denoted by $\Lambda_0(C)$. Let $C_1$ be a set of
constants typed by types from $\mbox{\emph{TYP}}(\mathcal{B}_1)$ and
let $C_2$ be a set of constants typed by types from
$\mbox{\emph{TYP}}(\mathcal{B}_2)$. A function
$\vartheta:\Lambda(C_1)\to\Lambda(C_2)$ is a \emph{term homomorphism
based on $\eta$} if
$\eta:\mbox{\emph{TYP}}(\mathcal{B}_1)\to\mbox{\emph{TYP}}(\mathcal{B}_2)$
is a type homomorphism and, whenever $M\in\Lambda(C_1)$:
\begin{itemize}
\item $\vartheta(M)$
is a term of type $\eta(\tau)$, if $M$ is a constant of type $\tau$;
\item $\vartheta(M)$ is the $n$-th variable of type $\eta(\tau)$, 
if $M$ is the $n$-th variable of type $\tau$;
\item $\vartheta(M)=(\vartheta(A)\vartheta(B))$, if $M\equiv(AB)$;
\item $\vartheta(M)=\lambda y.\vartheta(A)$, where
$y=\vartheta(x)$, if $M\equiv(\lambda x.A)$.
\end{itemize}
Note that this implies that $\vartheta(M)$ is a term of type
$\eta(\tau)$, if $M$ is of type $\tau$.

Clearly, a term homomorphism $\vartheta$ with domain $\Lambda(C)$ is
completely determined by the values $\vartheta(c)$ for $c\in C$. This
continues to hold if we restrict the domain to the set of linear
lambda terms $\Lambda_0(C)$.

In order to show how this can be used, let us continue the example we
just looked at. Consider the (abstract) constants in the first column
of Table \ref{tab}, typed by the (abstract) types in the second
column. We can now define a term homomorphism $H_0$ by sending the
constants in the first column to their images in the third column,
making sure that these have types as in the fourth. Since $H_0$ is
assumed to be a type homomorphism, \emph{all} lambda terms over the
constants in the first column will now automatically have images under
$H_0$. For example, it can be easily checked that $H_0$ sends the
abstract term\footnote{We use the standard notation of lambda terms.
The application of $M$ to $N$ is written as $(MN)$ (not as $M(N)$) and
lambda abstractions are of the form $(\lambda X.A)$. The usual
redundancy rules for parentheses apply, but will often not be used in
abstract terms, in order to emphasise their closeness to linguistic
expressions. In some cases, where it seems to improve clarity, we will
flout the rules and write things like $M(N_1,\ldots, N_n)$ for
$(MN_1\ldots N_n)$ or $A\land B$ for $\mathord{\land}AB$.}
\[((\mbox{\tt a
woman})\lambda \xi((\mbox{\tt every man})( \mbox{\tt loves }\xi)))\]
(in which $\xi$ is of type $D$), to a term $\beta\eta$-equivalent with
\[\lambda w\exists y({\it woman}\, yw\land \forall x({\it man}\,
xw\to {\it love}\, yxw))\ .\]
This term denotes the set of worlds in which some specific woman is
loved by all men. 

While this example sends abstract terms to translations that are close
to the ones of \cite{mont:prop74} and while such translations
obviously will not do as a \emph{vector} semantics, we will show in
the next sections that it is possible to alter the object language
while retaining the general translation mechanism. For more
information about the procedure of obtaining an object language from
an abstract language, see the papers mentioned or the explanation in
\cite{newdirections1}.

\section{A Static Vector Semantics}

\subsection{Vector Interpretations for the Object Language}
In order to provide an interpretation of our object language, the type
theory used must be able to talk about vectors over some field, for
which we choose the reals. We need a basic object type $R$ such that,
in all interpretations under consideration, the domain $D_R$ of type
$R$ is equal to or `close enough' to the set of reals $\mathbb{R}$ and
such that constants such as the following have their usual
interpretation (the $\to$ in types is dropped and association is to
the right; constants such as $+$, $\cdot$, and $<$ will be written
between their arguments). 
\begin{eqnarray*}
0&:& R\\
 1&:&R\\
 +&:&RRR\\
  \cdot&:&RRR\\
  {<}&:& RRR
  \end{eqnarray*} 
  
This can be done by
imposing one of the sets of second-order axioms in
\cite{tarski:intro}. Given these axioms we have that $D_R=\mathbb{R}$ in full
models, while we get non-standard models under the Henkin
interpretation.

Vectors can now be introduced as objects of type $IR$, where $I$ is
interpreted as some finite index set. Think of $I$ as a set of words;
if a word is associated with a vector $v:IR$, $v$ assigns a real  number to
each word, which gives information about the company the word keeps.
Since $IR$ will be
used often, we will abbreviate it as $V$. Similarly,  $IIR$, abbreviated
as $M$, can be associated with the type of \emph{matrices} and $IIIR$,
abbreviated as $C$, with the type of \emph{cubes}, and so on, see Table \ref{tb:TypesAbb}. 

\begin{table}
\begin{center}
\begin{tabular}{c|cc}
Type \quad & \quad Abbreviation &   \\
\hline
$IR$ & $V$ & Vector\\
$IIR$ & $M$ & Matrix\\
$IIIR$ & $C$ & Cube\\
$IIIIR$ & $H$ & Hypercube\\
$\cdots$&&
\end{tabular}
\end{center}
\caption{Vector types and their abbreviations}
\label{tb:TypesAbb}
\end{table}

\noindent In this paper  we will work with a single index type, but in general one can also consider cases with 
several index types, so that phrases of distinct
categories can live in their own space.

We need a toolkit of functions combining vectors, matrices, cubes,
etc. Here are some definitions, in which $r$ is of type $R$; $i$, $j$,
and $k$ are of type $I$; $v$ and $u$ are of type $V$; and $m$ and $c$
are of types $M$ and $C$ respectively. Indices are written as
subscripts---$v_i$ is syntactic sugar for $vi$.
\begin{align*}
* &:= \lambda rvi. r\cdot v_i:RVV\\
\boxplus &:= \lambda vui.v_i+u_i:VVV\\
\odot &:=\lambda vui.v_i\cdot u_i:VVV\\
\times_1 &:=\lambda mvi.\sum_j m_{ij}\cdot v_j:MVV\\
\times_2 &:=\lambda cvij.\sum_k m_{ijk}\cdot v_k:CVM\\
\langle\cdot\mid\cdot\rangle &:=\lambda uv.\sum_iu_i+v_i:VVR
\end{align*}

\noindent The reader will recognise $*$ as scalar product, $\boxplus$
as pointwise addition, $\odot$ as pointwise multiplication, $\times_1$
and $\times_2$ as matrix-vector and cube-vector multiplication, and
$\langle\cdot\mid\cdot\rangle$ as the dot product. One can also 
consider further operations, such as a \emph{rotation} operation $\rho:VVV$, given below
\[
\rho (\cdot, \cdot) :=   \lambda vu. 
\left ( 
\begin{array}{cc}
\langle u \mid v\rangle & \pm\sqrt{1- \langle u \mid v\rangle^2} \\
&\\
 \pm\sqrt{1- \langle u \mid v\rangle^2} &\langle u \mid v\rangle
 \end{array} \right ) 
 \times \frac{u + v}{2}
 \]
This operation takes two vectors and produces the result of rotating
their average towards the first vector. This can be thought of as, for
example, providing a good candidate for modelling head-argument
combinations. We will not present a specific use for this operation in
this paper and are just introducing it as an example of a new
operation that can be, but has not been, used by the existing vector
models, in order to illustrate the range of vector operation that can
be modelled in our setting. 

\subsection{Abstract Types and Type and Term Homomorphisms}
Let us assume again that our basic abstract types are $D$ for
determiner phrases, $S$ for sentences, and $N$ for nominal phrases.
But this time our type and term homomorphisms will be chosen in a way
different from how it was done in section 2. A very simple type
homomorphism $h$ can be defined by letting 
\[h(D)=h(S)=h(N)=V \]
So $h$ assigns vectors to determiners, nominal phrases and sentences.
There are other possibilities for the range of $h$ and we will sketch
a more elaborate assignment in which a running context is used in the
next section.  The above simple $h$ is chosen for the expository
purposes of this section. 

In Table \ref{tb:matrix-types}, we again provide abstract constants in
the first column and their abstract types in the second column; $h$
assigns to these the object types in the fourth column.  For instance,
the constant \texttt{woman} has the abstract type $N$, and a term
homomorphic image \textsf{woman}, which is assigned the type $V$ by
$h$. We say that the translation of \texttt{woman} is of type $V$. 
Similarly, the translations of \texttt{tall} and \texttt{smoke} are of
type $VV$, \texttt{love} and \texttt{know} are of type $VVV$, and
those of \texttt{every}, and \texttt{a} are of type $VV$.  The term
homomorphism $H$ is defined by letting its value for any abstract
constant in the first column be the corresponding object term in the
third column. Using this table, we automatically obtain homomorphic
images of any lambda term over the constants. But now our previous
example term\footnote{The entry for ${\tt man}$ is no longer present
in Table \ref{tb:matrix-types}. But ${\tt man}$ can be treated in full
analogy to ${\tt woman}$. In further examples we will also use
constants whose entries can be easily guessed.}
\[
((\mbox{\tt a woman})\lambda \xi((\mbox{\tt every
man})( \mbox{\tt loves }\xi)))
\]
is sent to a term that is $\beta\eta$
equivalent with \[(\mbox{\textsf{love}}\times_2
(\mbox{\textsf{a}}\times_1
\mbox{\textsf{woman}}))\times_1 (\mbox{\textsf{every}}\times_1
\mbox{\textsf{man}})\ .\] 

\begin{table}[t]
\begin{center}
{\small\def\arraystretch{1.4}\tabcolsep=5pt
\begin{tabular}{llll}\hline
$c$&$\tau$&
$H(c)$&$h(\tau)$\\ \hline
\texttt{woman}&$N$& \textsf{woman}&$V$\\
\texttt{tall}&$NN$&$\lambda v.(\mbox{\sf tall}\times_1 v)$&$VV$ \\
\texttt{smokes}&$DS$&$\lambda v.(\mbox{\sf smoke}\times_1 v)$&
$VV$\\
\texttt{loves}&$DDS$&$\lambda uv.
(\mbox{\sf love}\times_2 u)\times_1 v$&$VVV$\\
\texttt{knows}&$SDS$&$\lambda uv.
(\mbox{\sf know}\times_2 u)\times_1 v$&$VVV$\\
\texttt{every}&$N(DS)S$&$\lambda vZ.
Z(\mbox{\textsf{every}}\times_1 v)$&$V(VV)V$\\
\texttt{a}&$N(DS)S$&$\lambda vZ.
Z(\mbox{\textsf{a}}\times_1 v)$&$V(VV)V$\\
\hline
\end{tabular}
} \caption{Some abstract constants $c$ typed with abstract types
$\tau$ and their term homomorphic images $H(c)$ typed by $h(\tau)$.
Here, $Z$ is a variable of type $VV$, and $v$ and $u$ are of type
$V$.} \label{tb:matrix-types}
\end{center}
\end{table}


Nominal phrases like {\sf woman} are represented by vectors in Table
\ref{tb:matrix-types}, adjectives and intransitive verbs like {\sf
tall} and {\sf smoke} by matrices, and transitive verbs ({\sf love})
by cubes, as are constants like {\sf know}. Generalised quantifiers
are functions that take vectors to vectors. The composition operations
used ($\times_1$ and $\times_2$) are cube-vector and matrix-vector
instances of tensor contraction. The jury is still very much out on
what are the best operations for composing vectors.
\cite{Mitchell-Lapata} consider pointwise addition and multiplication
of vectors, matrix multiplication is used in \cite{BaroniZam}.  Such
operations are available to our theory. The table for these will have
a different $H(c)$ column and will be the same in all the other
columns. The $H(c)$ columns for these models are given in Table
\ref{tb:add-mult}.

\noindent
\begin{table}[t]
\begin{center}
{\small\def\arraystretch{1.4}\tabcolsep=5pt
\begin{tabular}{c|c|c}
\hline
Addition & Multiplication & Matrix Multiplication\\
\hline \hline
$H(c)$ & H(c) & H(c)\\ 
\hline
${\sf woman}$ & ${\sf woman}$ & ${\sf woman}$\\
$\lambda v.(\mbox{\sf tall}\boxplus v)$ & $\lambda v.(\mbox{\sf tall}\odot v)$  & $\lambda v.(\mbox{\sf tall}\times_1 v)$\\
$\lambda v.(\mbox{\sf smoke}\boxplus v)$ & $\lambda v.(\mbox{\sf smoke}\odot v)$ & $\lambda v.(\mbox{\sf smoke}\times_1 v)$\\
$\lambda uv.
(\mbox{\sf love}\boxplus u)\boxplus v$ & $\lambda uv.
(\mbox{\sf love}\odot u)\odot v$ & $\lambda uv.
(\mbox{\sf love}\times_1 u)\times_1 v$\\
$\lambda uv.(\mbox{\sf know}\boxplus u)\boxplus v$ & $\lambda uv.(\mbox{\sf know}\odot u)\odot v$ & $\lambda uv.(\mbox{\sf know}\times_1 u)\times_1 v$\\
$\lambda vZ.Z(\mbox{\textsf{every}}\boxplus v)$ & $\lambda vZ.Z(\mbox{\textsf{every}}\odot v)$ & $\lambda vZ.Z(\mbox{\textsf{every}}\times_1 v)$\\
$\lambda vZ.
Z(\mbox{\textsf{a}}\boxplus v)$ & $\lambda vZ.
Z(\mbox{\textsf{a}}\odot v)$ & $\lambda vZ.
Z(\mbox{\textsf{a}}\times_1 v)$\\
\hline
\end{tabular}
}
\caption{Term homomorphic images $H(a)$ for pointwise addition and multiplication, and matrix multiplication as composition operations. In the case of addition and multiplication,  $Z$ is a variable of type $V$, for matrix multiplication it is of type $VV$; whereas $v$  and $c$ are of type $V$ for all three operations.}
\label{tb:add-mult}
\end{center}
\end{table}
In this paper, we will not choose between these operations. Instead of this we will
explore the question how to combine such functions once an initial set
of them has been established (and validated empirically). Functions in
the initial set will typically combine vector meanings of adjacent
phrases. Our aim, like the one in \cite{baroni-etal-2014} (who also
give an excellent introduction to and survey of the work that has been
done in compositional vector semantics), has been to give a general theory
that also includes dependencies between phrases that are not adjacent,
such as in topicalisation and relative clause formation.

\section{Dynamic Vector Semantics with Context Change Potentials}

\subsection{Heim's Files and Distributional Contexts}
Heim describes her contexts as files that have some kind of
information written on (or in) them. Context changes are operations
that update these files, e.g.\ by adding or deleting information from
the files.  Formally, a context is taken to be a set of sequence-world
pairs, in which the sequences come from some domain $\mathcal{D}_I$ of
individuals, as follows:
\[ ctx\subseteq \{(g,w) \mid g \colon
\mathbb{N} \to {\cal D}_I, w \ \mbox{a possible world}\} \]
%
%
%
%
We follow Heim \cite{Heim1983} here in letting the sequences in her
sequence-world-pairs be infinite, although they are best thought of as
finite.

Sentence meanings are \emph{context change potentials} (CCPs) in
Heim's work, functions from contexts to contexts. A sentence $S$ comes
provided with a sequence of instructions that, given any context
$ctx$, updates its information so that a new context denoted as
\[ ctx + S \] 
results. The sequence of instructions that brings about this update is
derived compositionally from the constituents of $S$.

In distributional semantics, contexts are words somehow related to
each other via their patterns of use, e.g.\  by co-occurring in a
neighbourhood word window of a fixed size or via a dependency
relation. In practice, one builds a context matrix $M$ over
$\mathbb{R}^2$, with rows and columns labeled by words from a
vocabulary $\Sigma$ and with entries taking values from $\mathbb{R}$,
for a full description see (\cite{Rubenstein1965}).  $M$ can be seen
as the set of its vectors: 
\[ \{\overrightarrow{v} \mid
\overrightarrow{v} \colon 
\Sigma \to \mathbb{R}\} \]
where each $\overrightarrow{v}$ is a row or column in $M$.

If we take Heim's domain of individuals ${\cal D}_I$ be the vocabulary
of a distributional model of meaning, that is ${\cal D}_I := \Sigma$,
then a context matrix can be seen as a so-called \emph{quantized}
version of a Heim context:
\[
 \{(\overrightarrow{g},w) \mid \overrightarrow{g} \colon 
 \Sigma \to \mathbb{R}, w \ \mbox{a possible world}\}
\]
Thus a distributional context matrix is obtainable by endowing Heim's
contexts with $\mathbb{R}$. In other words, we are assuming that not
only a file has a set of individuals, but also that these individuals take
some kind of values, e.g. from  reals.  

The role of possible worlds in a distributional semantics is arguable,
as vectors retrieved from a corpus are not naturally truth
conditional. Keeping the possible worlds in the picture provides a
machinery to assign a proposition to a distributional vector by other
means and can become very useful. We leave working with possible worlds to future work and in this paper only work with the
 sets of vectors as our contexts, that is:
\begin{equation}
\label{eq:vectorcontrext}
 ctx \subseteq \{\overrightarrow{g} \mid \overrightarrow{g} \colon 
 \Sigma \to \mathbb{R}, \ov{g} \in M\}
\end{equation}
Distributional versions of Heim's CCP's can be defined based on the
intuitions and definitions of Heim.  In what follows we pan out how
these instructions let contexts thread through vectorial semantics in
a compositional manner.

%

\subsection{Dynamic Type and Term Homomorphisms and their Interpretations}



On the  set of basic abstract types $D,S,N$ a \emph{dynamic} type homomorphism $\rho$  that takes into account the contexts of words is defined as follows:
\[
\rho(N)=(VU)U, \quad \rho(D)=V, \quad \rho(S)=U
\]
Here, sentences are treated as  \emph{context change potentials}. They update contexts and we assign the type $U$  (for `update') to them. A context can be a matrix or a cube, so it can have  type $I^2R$ or $I^3 R$. A sentence can then have  type
$(I^2R)(I^2R)$ or $(I^3R)(I^3R)$. We have previously abbreviated $IR$ to $V$, $I^2 R$ to $M$, and $I^3R$ to $C$. The sentence type then becomes $MM$  or $CC$,  used as an abbreviation for $U$; the former in the context matrix setting and the latter in the context cube setting.  The concrete semantics obtained by instantiating $U$ to matrix and cube contexts are discussed in more details in  sections \ref{sec:contextmatrix} and \ref{sec:contextcube},  respectively.

The update functions are presented in Table \ref{table:matrix-types}.
Simple words such as names, nouns, adjectives, and verbs are first
assigned vectors, denoted by constants such as
$\ov{\mbox{\textsf{anna}}}$, $\ov{\mbox{\textsf{woman}}}$,
$\ov{\mbox{\textsf{tall}}}$ and $\ov{\mbox{\textsf{smoke}}}$ (all of
type $V$).  These
are then used by the typed lambda calculus given via $H(a)$, in the
third column, to build certain functions, which will act as the
meanings of those words in context.  The object types assigned by
$\rho$ are as follows: 

\begin{center}
$\begin{array}{lcl}
\mbox{Type of nouns}&: & (VU)U\\
\mbox{Type of adjectives}&: & ((VU)U)(VU)U\\
\mbox{Type of intransitive verbs}&: &  VU\\
\mbox{Type of transitive verbs}&: &  VVU
\end{array}$
\end{center}

The function $Z$ updates the context of proper names and nouns based
on their vectors e.g. $\ov{\mbox{\textsf{anna}}}$ and
$\ov{\mbox{\textsf{woman}}}$. These are essentially treated as vectors
with type $V$, but, since they must be made capable of dynamic
behaviour, they are `lifted' to the higher type $(VU)U$.  

The function $F$ of an adjective, takes a vector for the adjective,
e.g. $\ov{\mbox{\textsf{tall}}}$, a vector for its argument, e.g. $v$
and a vector for its context, e.g. $c$, then updates the context, e.g.
as in $F(\ov{\mbox{\textsf{tall}}}, v,c)$.  The output of this
function is then lifted to the the higher type, i.e. $((VU)U)((VU)U)$
via the functions $Z$ and $Q$, respectively. 

Functions $G$ and $I$ update contexts of verbs; they take a vector for
the verb as well as a vector for each of its arguments, plus an input
context, and then return a context as their output. So, the function
$G$ of an intransitive verb takes a vector for itself, e.g.
$\ov{\mbox{\textsf{smoke}}}$ a vector for its subject, e.g. $v$, plus
a context, e.g. $c$, and returns a modified context, e.g. via $\lambda
vc. G(\ov{\mbox{\textsf{smoke}}}, v,c)$. The function $I$ of a
transitive verb takes a vector for itself, a vector for its subject, a
vector for its object and a context, and returns a context.

\noindent
\begin{table}[t]
\begin{center}
{\small\def\arraystretch{1.4}\tabcolsep=5pt
\begin{tabular}{llll}\hline
$a$&$\tau$&
$H(a)$&$\rho(\tau)$\\ \hline
\texttt{Anna}&$(DS)S$&$\lambda Z.Z (\ov{\mbox{\textsf{anna}}})$&$(VU)U$\\
\texttt{woman}&$N$& $\lambda Z.Z (\ov{\mbox{\textsf{woman}}})$&$(VU)U$\\
\texttt{tall}&$NN$&$\lambda QZ.Q(\lambda vc.ZvF(\ov{\mbox{\sf tall}},v,
c))$&$((VU)U)(VU)U$\\
\texttt{smokes}&$DS$&$\lambda vc.G(\ov{\mbox{\sf smoke}}, v,c)$&
$VU$\\
\texttt{loves}&$DDS$&$\lambda uvc.
I(\ov{\mbox{\sf love}},u,v,c) $&$VVU$\\
\texttt{knows}&$SDS$&$\lambda pvc.pJ(\ov{\mbox{\sf know}},v,c)$&$UVU$\\
\texttt{every}&$N(DS)S$&$\lambda Q.Q$&$((VU)U)(VU)U$\\
\texttt{who}&$(DS)NN$&$\lambda Z'QZ.Q(\lambda vc.Zv(QZ'c))$&
$(VU)((VU)U)(VU)U$\\
\texttt{and}& {\small ${(\ovl{\alpha}S)(\ovl{\alpha}S)(\ovl{\alpha}S)}$}&
$\lambda R'\lambda R\lambda \ovl{X}\lambda c.R'\ovl{X}(R\ovl{X}c)$&
$(\rho(\ovl{\alpha})U)(\rho(\ovl{\alpha})U)(\rho(\ovl{\alpha})U)$\\
\hline\\
\end{tabular}
} 
\caption{Some abstract constants $a$ typed with abstract types
$\tau$ and their term homomorphic images $H(a)$ typed by $\rho(\tau)$
(where $\rho$ is a type homomorphism, i.e.\
$\rho(AB)=\rho(A)\rho(B)$). Here $Z$ is a variable of type $VU$, $Q$
is of type $(VU)U$, $v$ of type $V$, $c$ of type $M$, and $p$ and $q$
are of type $U$. The functions $F$, $G$, $I$, and $J$ are explained in
the main text. In the schematic entry for \texttt{and}, we write 
$\rho(\ovl{\alpha})$ for $\rho(\alpha_1)\cdots\rho(\alpha_n)$, if
$\ovl{\alpha}=\alpha_1\cdots\alpha_n$.}
\label{table:matrix-types}
\end{center}
\end{table}

The meanings of function words, such as conjunctions, relative
pronouns, and quantifiers, will not (necessarily) be identified with
vectors.  The type of the quantifier \emph{every} is $((VU)U)(VU)U$,
where its noun argument has the required `quantifier' type $(VU)U$.
The lambda calculus entry for `every', $\lambda Q.Q$, is the identity
function; it takes a $Q$ and then spits it out again. The alternative
would be to have an entry along the lines of that of `tall', but this
would not make a lot of sense. It is the content words that seem to be
important in a distributional setting, not the function words.

The word \emph{and} is treated as a generalised form of function
composition. Its entry is schematic, as \emph{and} does not only
conjoin sentences, but also other phrases of any category. So, the
type of the abstract constant connected with the word is
${(\ovl{\alpha}S)(\ovl{\alpha}S)(\ovl{\alpha}S)}$, in which
$\ovl{\alpha}$ can be any sequence of abstract types. Ignoring this
generalisation for the moment, we obtain $SSS$ as the abstract type
for sentence conjunction, with a corresponding object type $UUU$, and
meaning $\lambda pqc.p(qc)$, which is just function composition. This
is defined in a way such that the context updated by \emph{and}'s left
argument will be further updated by its right argument. So `Sally
smokes and John eats bananas' will, given an initial context $c$,
first update $c$ to $G(\mbox{\sf Sally}, \mbox{\sf smoke}, c)$, which
is a context, and then update further with `John eats bananas' to
$I(\mbox{\sf eat}, \mbox{\sf John}, \mbox{\sf bananas}, G(\mbox{\sf
smoke}, \mbox{\sf Sally}, c))$.  This treatment of \emph{and} is
easily extended to coordination in all categories. For example, the
reader may check that \texttt{and admires loves} (which corresponds to
\emph{loves and admires}) has $\lambda uvc.I(\ov{\mbox{\sf
admire}},u,v,I(\ov{\mbox{\sf love}},u,v,c))$ as its homomorphic image.

The update instructions fall through the semantics of phrases and
sentences compositionally.  The sentence \emph{every tall
woman smokes}, for example, will be associated 
with the following  lambda expression:
\[
(\mbox{\tt ((every (tall woman)) smokes)})
\]
 This in
its turn has a term homomorphic image that is 
$\beta$-equivalent with the following:
\[\lambda c.G\left(\ov{\mbox{\sf smoke}}, \ov{\mbox{\sf
woman}},F(\ov{\mbox{\sf tall}}, \ov{\mbox{\sf woman}},c)\right)\]
which describes a distributional context update for it.  This term
describes a first update of the context $c$ according to the rule for
the constant {\tt tall}, and then a second update according to the
rule for the constant {\tt smokes}. As a result of these, the value
entries at the crossings of $\langle${\sf tall}, {\sf woman}$\rangle$
and $\langle${\sf woman}, {\sf smokes}$\rangle$ get increased. Much
longer chains of context updates can be `threaded' in this way.

In the following,  we give some examples. In each case the a.\ sentence
is followed by an abstract term in b.\ which captures its syntactic
structure. The update potential that follows in c.\ is the homomorphic
image of this abstract term.

\eenumsentence{
\item Sue loves and admires a stockbroker
\item $\mbox{\texttt{(a stockbroker)}}\lambda \xi.
\mbox{\texttt{Sue(and admires loves $\xi$)}}$
\item $\lambda c.I(\ov{\mbox{\sf admire}},\ov{\mbox{\sf stockbroker}},
\ov{\mbox{\sf sue}},I(\ov{\mbox{\sf love}},\ov{\mbox{\sf stockbroker}},
\ov{\mbox{\sf sue}},c))$
}

\eenumsentence{
\item Bill admires but Anna despises every cop
\item \texttt{(every cop)}$\lambda\xi.$%
\texttt{and(Anna(despise $\xi$))(Bill(admire $\xi$))}
\item $\lambda c.I(\ov{\mbox{\sf despise}},\ov{\mbox{\sf cop}},
\ov{\mbox{\sf anna}},I(\ov{\mbox{\sf admire}},\ov{\mbox{\sf cop}},
\ov{\mbox{\sf bill}},c))$
}

\eenumsentence{
\item The witch who Bill claims Anna saw disappeared
\item \texttt{the(who($\lambda\xi.$Bill(claims(Anna(saw
$\xi$))))witch)disappears}
\item $\lambda c.G(\ov{\mbox{\sf disappear}},\ov{\mbox{\sf witch}},
I(\ov{\mbox{\sf see}},\ov{\mbox{\sf witch}},
\ov{\mbox{\sf anna}},J(\ov{\mbox{\sf claim}},\ov{\mbox{\sf bill}},c)))$
}

%


\section{Co-occurence Matrix Context and its Update}
\label{sec:contextmatrix}

In this section we assume that our contexts are the co-occurrence matrices of distributional semantics \cite{Rubenstein1965}.  Given a corpus of text, a co-occurrence  matrix has at each of its entries   the degree of co-occurrence between a  word and its neighbouring words.  The neighbourhood is usually a window of  $k$ words to both sides of the word. The update type $U$ associated to sentences,  will thus take the form $(I^2R)(I^2R)$, abbreviated to $MM$.  That is, a sentence will take a co-occurrence matrix as input, update it with new entries, and return the updated matrix as output.  

Since we are working with co-occurrence matrices, the updates simply increase the degrees of co-occurrences between the labelling words of the rows and columns of the matrix. In this  paper, for the purpose of keeping it simple,  we work with a  co-occurrence matrix with raw co-occurrence numbers as entries.  In this case, the update functions just add 1 to each entry  at each single update step. This may be extended to (or replaced with)  logarithmic probabilistic entries  such as Pointwise Mutual Information (PMI) or its positive or smoothed version PPMI, PPMI$_\alpha$,  in which case the update functions have to recalculate these weighting schemes at each step.   For instance, see the example presented in Table \ref{table:example-matrix}. The cells whose entries are increased are chosen according  to the grammatical roles  of the labelling words. These are implemented in the functions $F, G, I, J$, which   apply the updates to each word of the sentence. The updates are compositional, i.e. they  can be applied  compositionally to the words within a sentence. This is evident  as the updates induced by words of a sentence are designed based on  the grammatical roles of them and this acts a glue. 

More formally, the  object terms corresponding to a  word $a$  update
a context matrix $c$ with the information in $\ov{a}$ and the
information in the vectors of arguments $u,v, \cdots$ of $a$. The result is a new
context matrix $c'$, with different value entries, depicted below:
{\small
\[
\left ( \begin{array}{ccc}
m_{11} & \cdots & m_{1k}\\
m_{21} &  \cdots & m_{2k}\\
\vdots&&\\
m_{n1} &  \cdots & m_{nk}
\end{array} \right) 
 +  \langle \ov{a}, u,v, \cdots \rangle = 
\left ( \begin{array}{ccc}
m'_{11} & \cdots & m'_{1k}\\
m'_{21} &  \cdots & m'_{2k}\\
\vdots&&\\
m'_{n1} &  \cdots & m'_{nk}
\end{array} \right)
\]
}
where we have
\[
m'_{ij} := m_{ij} + 1
\]
The $m_{ij}$ and $m'_{ij}$ entries are described as follows:
%
\begin{itemize}
\item The function denoted by $\lambda vc.G(\ov{\mbox{\sf smoke}}, v, c)$ 
increases the value entry of $m_{ij}$ of $c$ by 1, for $i$ and $j$ indices
of {\sf smoke} and its subject $v$. 

\item The function denoted by $\lambda uv.\lambda c.I(\ov{\mbox{\sf love}}, u,v, c)$ 
increases the value entries of  $m_{ij}$, $m_{jk}$, and $m_{ik}$ of $c$ by 1, for
$i,j,k$ indices of {\sf loves}, its subject $u$ and its object $v$. 
\item The function denoted by $\lambda vc.F(\ov{\mbox{\sf tall}}, v, c)$ 
increases the value entry of $m_{ij}$ of $c$ by 1, for $i$ and $j$ indices
of {\sf tall} and its modified noun $v$. The entry for \emph{tall} in
Table 1 uses this function, but allows for further update of context.
\item The function denoted by $\lambda vc.J(\ov{\mbox{\sf know}}, v, c)$
increases the value entry of $m_{ij}$ of $c$ by 1, for $i$ and $j$ indices
of {\sf know} and its subject $v$. The updated matrix is made the
input for further update (by the context change potential of the
sentence that is known) in Table 1.
\end{itemize}

\begin{figure}[b!]
\hspace{-0.1cm}
\begin{tabular}{cl|ccccc}
&& 1&2&3&4&5 \\
&&man&  cat  & loves & fears & sleeps\\
\hline
&\\
1& Anna& 100 &700& 800&  500 & 400 \\
&\\
2& woman& 500 & 650& 750& 750 &  600 \\
&\\
3& tall &  300& 50& 500& 400 & 400\\
&\\
4& smokes&  400&50 &600& 600 & 200 \\
&\\
5& loves &  350&  250 & $\epsilon$ & 600 & 500 \\
&\\
6& knows& 300  & 50& 200& 250  & 270\\
&\\
\end{tabular}
\qquad $\stackrel{\stackrel{\text{update \ by}}{\implies}}{\small G,I,F,J}$
\qquad
\begin{tabular}{cl|ccccc}
&& 1&2&3&4&5 \\
&&man&  cat  & loves & fears & sleeps\\
\hline
&\\
1& Anna& 100 &700& 800&  500 & 400 \\
&\\
2& woman& 500 & 650& 750& 750 &  600 \\
&\\
3& tall &  650& 50& 500& 400 & 400\\
&\\
4& smokes&  700&50 &600& 600 & 200 \\
&\\
5& loves &  550&  750 & $\epsilon$ & 600 & 500 \\
&\\
6& knows& 600  & 250& 450& 510  & 700\\
&\\
\end{tabular}
\caption{An example of updates by functions $F,G,I,J$ on a co-occurrence matrix}
\label{table:matrices}
\end{figure}

As an example, consider the  co-occurrence matrices  depicted in  Figure \ref{table:matrices}.  The left hand side matrix is a snapshot of a matrix just before  a series of updates are applied to it.  In this matrix, for individual nouns such as \texttt{Anna} and noun phrases such as \text{woman} we assume $Z$ is the identity. So their entries are the same as the entries of their distributional co-occurrence vectors. The rationale of this example for the entries of these words is as follows: \texttt{Anna} is a woman and so it has not much appeared  in a corpus in the neighbourhood of  the context \texttt{men}; as a result,  it has a low value of 100 at that entry; she loves cats (and has some herself), so her entry at the context \texttt{cat} is 700; she loves other things such as smoking, and  so she has a substantial entry at the context \texttt{loves}; and so on.  The entries of  the other words, i.e. \texttt{tall, smokes, loves, knows}, are also initialised to their distributional co-occurrence matrix vectors.   When an entry $c_{ij}$ corresponds to the same two words, e.g. when $i$ and $j$ are both \texttt{love} as in the left hand side matrix of Figure  \ref{table:matrices}, we put a $\epsilon$ to indicate a predefined fixed value. 

The intransitive verb \texttt{smokes} updates the left hand side  matrix  of Figure  \ref{table:matrices} via the function $G$ at the entries $c_{4j}$. Here,  in principle, $j$ can be 1 and 2, as both \texttt{man} and \texttt{cat}, in their singular or plural forms, could have occurred as subjects of \texttt{smokes} in the corpus. Assuming that cats do not smoke and that a reasonable number of men do, a series of, for instance, 300 occurrences of  \texttt{smokes} with the subject \texttt{man}, updates this entry and raises  its value from 400 to 700. Similarly, adjective \texttt{tall} updates the entries of the $c_{3j}$ cells of the matrix via the function $F$, where $j$ can in principle be 1 and 2, but since cats are not usually tall, it only updates $c_{31}$. Again, a series of, for example,  350 occurrences of the adjective \texttt{tall} as the modifier of \texttt{man} raises this number  from 300 to, say 650. The case for \texttt{loves}  and function $I$ is similar. For \texttt{knows}, men know cats love mice, and love to play and be stroked,  etc; they know that cats fear water and objects such as  vacuum cleaners, and that they sleep a lot. As a result, the values of all of the entires of row 6, that is $c_{61}, c_{62}, c_{63}, c_{64}$ and $c_{65}$  will be updates by function $J$, for instance, to the  numbers in the right hand side table.

\section{Entity Relation Cube  Context and its Update}
\label{sec:contextcube}

\begin{figure}[b!]
\begin{center}
{%
\beginpgfgraphicnamed{cube1}
\begin{tikzpicture}
	\begin{pgfonlayer}{nodelayer}
		\node [style=none] (0) at (-5, 3) {};
		\node [style=none] (1) at (-5, -1) {};
		\node [style=none] (2) at (0, -1) {};
		\node [style=none] (3) at (0, 3) {};
		\node [style=none] (4) at (-2.5, 0) {};
		\node [style=none] (5) at (-2.5, -1) {};
		\node [style=none] (6) at (-5, 0) {};
		\node [style=none] (7) at (-2.75, 0.75) {};
		\node [style=none] (8) at (-1, 0.5) {$c_{ijk}$};
		\node [style=none] (9) at (-5, 3.5) {entity};
		\node [style=none] (10) at (0.75, 3.5) {relation};
		\node [style=none] (11) at (1.25, -1) {entity};
	\end{pgfonlayer}
	\begin{pgfonlayer}{edgelayer}
		\draw [<-] (0.center) to (1.center);
		\draw [->] (1.center) to (2.center);
		\draw [->] (1.center) to (3.center);
		\draw [dashed] (6.center) to (4.center);
		\draw [dashed] (4.center) to (5.center);
		\draw [dashed] (7.center) to (4.center);
	\end{pgfonlayer}
\end{tikzpicture}}
\endpgfgraphicnamed} 
$ \quad + \quad (\ov{a}, u, v)  \quad = \quad $
{%
\beginpgfgraphicnamed{cube2}
\begin{tikzpicture}
	\begin{pgfonlayer}{nodelayer}
		\node [style=none] (0) at (-5, 3) {};
		\node [style=none] (1) at (-5, -1) {};
		\node [style=none] (2) at (0, -1) {};
		\node [style=none] (3) at (0, 3) {};
		\node [circle, fill, style=none] (4) at (-2, 2.5) {};
		\node [style=none] (5) at (-2, -1) {};
		\node [style=none] (6) at (-5, 2.5) {};
		\node [style=none] (7) at (-1, 2.25) {};
		\node [style=none] (8) at (-2, 3.5) {$c'_{ijk}$};
		\node [style=none] (9) at (-5, 3.5) {entity};
		\node [style=none] (10) at (1, 3.5) {relation};
		\node [style=none] (11) at (1.25, -1) {entity};
	\end{pgfonlayer}
	\begin{pgfonlayer}{edgelayer}
		\draw [<-] (0.center) to (1.center);
		\draw [->] (1.center) to (2.center);
		\draw [->] (1.center) to (3.center);
		\draw [dashed] (6.center) to (4.center);
		\draw [dashed] (4.center) to (5.center);
		\draw [dashed] (7.center) to (4.center);
	\end{pgfonlayer}
\end{tikzpicture}}
\endpgfgraphicnamed}

where 
\[
c'_{ijk} := c_{ijk} + 1
\]
\end{center}
\caption{Updates of entries in an entity relation cube}
\label{fig:cubeupdate}
\end{figure}
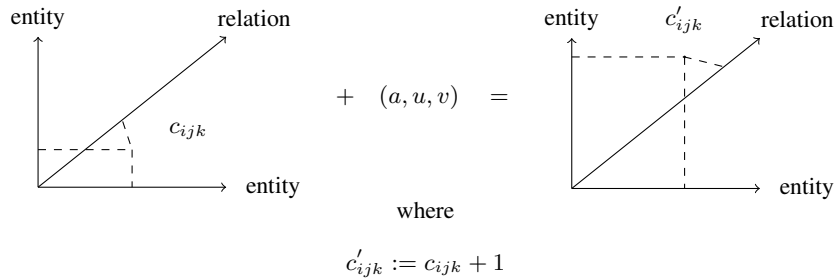

A  corpus of text can be seen as  a  sequence of lexical items occurring in the vicinities of each other and turned into a co-occurrence matrix. It can also be seen as a sequence of entities related to each other via predicate-argument structures and turned into an entity relation graph. These can be modelled in our setting by taking the contexts to be cubes,  thus setting $S$ to have  the update type $U = (I^3R)(I^3R)$, abbreviated to $CC$.     The entity relation graph  approach needs a more costly preprocessing of the corpus, but it is useful for a systematic treatment of   logical words such as negation and quantification, as well as coordination. 

An entity relation graph can be derived from  different   resources: a semantic network of concepts, a knowledge base such as WordNet or FrameNet,   or an olog or a Gellish network. We work with  entities and  relations   extracted from text. Discovering such graphs from corpora of text in an automatic way has been subject of much recent research, for example  see   \cite{Yao2012,Riedel2010} for a direct approach and   \cite{Kambhatla2004,Poon2009} via semantic parsing.  The elements of an entity relation graph are argument-relation-argument triples, sometimes referred to by \emph{relation paths} \cite{Yao2012}. Similar to  \cite{Lewis-Steedman2013} we base ourselves in a binary version of the world, where our relations are all binary; we  turn  non-binary relations into binary ones using the  {\sf is-a}  predicate.

Similar to the matrix case, the object terms corresponding to a constant $a$  update
a context cube $c$ with the information in $\ov{a}$ and the
information in the vectors of arguments of $a$. The result is a new
context cube $c'$, with different value entries, greater or less than the originals, as depicted in Figure \ref{fig:cubeupdate}.

The $c_{ijk}$ and $c'_{ijk}$ entries are   similar to the matrix case, for example 
\begin{itemize}
\item The function denoted by $\lambda vc.G(\ov{\mbox{\sf smoke}}, v, c)$ 
increases the value entry $c_{ijk}$ of $c$, for $i,j,k$  indices
of {\sf is-a} and {\sf smoker} and $v$  the subject of smoke.  

\item The function denoted by $\lambda vc.F(\ov{\mbox{\sf tall}}, v, c)$ 
increases the value entry $c_{ijk}$ of $c$, for $i,j,k$ indices
of {\sf is} and  {\sf tall} and its modified noun $v$. 

\item The function denoted by $\lambda uv c.I(\ov{\mbox{\sf love}}, u,v, c)$ 
increases the value entries of $c_{ijk}$ of $c$, for
$i,j,k$ indices of {\sf loves}, its subject $u$ and its object $v$. 
\end{itemize}

\begin{figure}[h]
\begin{center}
{%
\beginpgfgraphicnamed{cube1smoke}
\begin{tikzpicture}
	\begin{pgfonlayer}{nodelayer}
		\node [style=none] (0) at (-5, 3) {};
		\node [style=none] (1) at (-5, -1) {};
		\node [style=none] (2) at (0, -1) {};
		\node [style=none] (3) at (0, 3) {};
		\node [style=none] (4) at (-2.5, 0) {};
		\node [style=none] (5) at (-2.5, -1) {};
		\node [style=none] (6) at (-5, 0) {};
		\node [style=none] (7) at (-2.75, 0.75) {};
		\node [style=none] (8) at (-1.75, 0) {$100$};
		\node [style=none] (9) at (-5, 3.5) {Anna};
		\node [style=none] (10) at (0.75, 3.5) {\bf is-a};
		\node [style=none] (11) at (1.25, -1) {smoker};
	\end{pgfonlayer}
	\begin{pgfonlayer}{edgelayer}
		\draw [<-] (0.center) to (1.center);
		\draw [->] (1.center) to (2.center);
		\draw [->] (1.center) to (3.center);
		\draw [dashed] (6.center) to (4.center);
		\draw [dashed] (4.center) to (5.center);
		\draw [dashed] (7.center) to (4.center);
	\end{pgfonlayer}
\end{tikzpicture}}
\endpgfgraphicnamed} $\stackrel{\text{update \ by\ } G}{\implies}${%
\beginpgfgraphicnamed{cube2smokeU}
\begin{tikzpicture}
	\begin{pgfonlayer}{nodelayer}
		\node [style=none] (0) at (-5, 3) {};
		\node [style=none] (1) at (-5, -1) {};
		\node [style=none] (2) at (0, -1) {};
		\node [style=none] (3) at (0, 3) {};
		\node [circle, fill, style=none] (4) at (-2, 2.5) {};
		\node [style=none] (5) at (-2, -1) {};
		\node [style=none] (6) at (-5, 2.5) {};
		\node [style=none] (7) at (-1, 2.25) {};
		\node [style=none] (8) at (-2, 3) {$400$};
		\node [style=none] (9) at (-5, 3.5) {Anna};
		\node [style=none] (10) at (1, 3.5) {\bf is-a};
		\node [style=none] (11) at (1.25, -1) {smoker};
	\end{pgfonlayer}
	\begin{pgfonlayer}{edgelayer}
		\draw [<-] (0.center) to (1.center);
		\draw [->] (1.center) to (2.center);
		\draw [->] (1.center) to (3.center);
		\draw [dashed] (6.center) to (4.center);
		\draw [dashed] (4.center) to (5.center);
		\draw [dashed] (7.center) to (4.center);
	\end{pgfonlayer}
\end{tikzpicture}}
\endpgfgraphicnamed} 

\smallskip
{%
\beginpgfgraphicnamed{cube1tall}
\begin{tikzpicture}
	\begin{pgfonlayer}{nodelayer}
		\node [style=none] (0) at (-5, 3) {};
		\node [style=none] (1) at (-5, -1) {};
		\node [style=none] (2) at (0, -1) {};
		\node [style=none] (3) at (0, 3) {};
		\node [style=none] (4) at (-2.5, -0.5) {};
		\node [style=none] (5) at (-2.5, -1) {};
		\node [style=none] (6) at (-5, -0.5) {};
		\node [style=none] (7) at (-3.75, 0) {};
		\node [style=none] (8) at (-2, 0) {$50$};
		\node [style=none] (9) at (-5, 3.5) {Anna};
		\node [style=none] (10) at (0.5, 3.25) {\bf is};
		\node [style=none] (11) at (1.25, -1) {tall};
	\end{pgfonlayer}
	\begin{pgfonlayer}{edgelayer}
		\draw [<-] (0.center) to (1.center);
		\draw [->] (1.center) to (2.center);
		\draw [->] (1.center) to (3.center);
		\draw [dashed] (6.center) to (4.center);
		\draw [dashed] (4.center) to (5.center);
		\draw [dashed] (7.center) to (4.center);
	\end{pgfonlayer}
\end{tikzpicture}}
\endpgfgraphicnamed} $\stackrel{\text{update \ by\ } F}{\implies}${%
\beginpgfgraphicnamed{cube2tallU}
\begin{tikzpicture}
	\begin{pgfonlayer}{nodelayer}
		\node [style=none] (0) at (-5, 3) {};
		\node [style=none] (1) at (-5, -1) {};
		\node [style=none] (2) at (0, -1) {};
		\node [style=none] (3) at (0, 3) {};
		\node [circle, fill, style=none] (4) at (-2.5, 1.5) {};
		\node [style=none] (5) at (-2.5, -1) {};
		\node [style=none] (6) at (-5, 1.5) {};
		\node [style=none] (7) at (-1.5, 1.75) {};
		\node [style=none] (8) at (-2.5, 2) {$280$};
		\node [style=none] (9) at (-5, 3.5) {Anna};
		\node [style=none] (10) at (0.5, 3.25) {\bf is};
		\node [style=none] (11) at (1.25, -1) {tall};
	\end{pgfonlayer}
	\begin{pgfonlayer}{edgelayer}
		\draw [<-] (0.center) to (1.center);
		\draw [->] (1.center) to (2.center);
		\draw [->] (1.center) to (3.center);
		\draw [dashed] (6.center) to (4.center);
		\draw [dashed] (4.center) to (5.center);
		\draw [dashed] (7.center) to (4.center);
	\end{pgfonlayer}
\end{tikzpicture}}
\endpgfgraphicnamed} 

\smallskip
{%
\beginpgfgraphicnamed{cube1love}
\begin{tikzpicture}
	\begin{pgfonlayer}{nodelayer}
		\node [style=none] (0) at (-5, 3) {};
		\node [style=none] (1) at (-5, -1) {};
		\node [style=none] (2) at (0, -1) {};
		\node [style=none] (3) at (0, 3) {};
		\node [style=none] (4) at (-3.25, 1) {};
		\node [style=none] (5) at (-3.25, -1) {};
		\node [style=none] (6) at (-5, 1) {};
		\node [style=none] (7) at (-2.75, 0.75) {};
		\node [style=none] (8) at (-3.25, 1.5) {200};
		\node [style=none] (9) at (-5, 3.5) {Anna};
		\node [style=none] (10) at (0.5, 3.5) {loves};
		\node [style=none] (11) at (0.75, -1) {cat};
	\end{pgfonlayer}
	\begin{pgfonlayer}{edgelayer}
		\draw [<-] (0.center) to (1.center);
		\draw [->] (1.center) to (2.center);
		\draw [->] (1.center) to (3.center);
		\draw [dashed] (6.center) to (4.center);
		\draw [dashed] (4.center) to (5.center);
		\draw [dashed] (7.center) to (4.center);
	\end{pgfonlayer}
\end{tikzpicture}}
\endpgfgraphicnamed} $\stackrel{\text{update \ by\ } I}{\implies}${%
\beginpgfgraphicnamed{cube2loveU}
\begin{tikzpicture}
	\begin{pgfonlayer}{nodelayer}
		\node [style=none] (0) at (-5, 2.5) {Anna};
		\node [style=none] (1) at (-3, 1.5) {350};
		\node [style=none] (2) at (0.5, 2.5) {loves};
		\node [style=none] (3) at (-3.25, 1) {};
		\node [style=none] (4) at (-5, -2) {};
		\node [style=none] (5) at (-5, 1) {};
		\node [style=none] (6) at (-5, 2) {};
		\node [style=none] (7) at (0.75, -2) {cat};
		\node [style=none] (8) at (0, -2) {};
		\node [style=none] (9) at (0, 2) {};
		\node [style=none] (10) at (-3.25, -2) {};
		\node [style=none] (11) at (-2, 0.5) {};
	\end{pgfonlayer}
	\begin{pgfonlayer}{edgelayer}
		\draw [<-] (6.center) to (4.center);
		\draw [->] (4.center) to (8.center);
		\draw [->] (4.center) to (9.center);
		\draw [dashed] (5.center) to (3.center);
		\draw [dashed] (3.center) to (10.center);
		\draw [dashed] (11.center) to (3.center);
	\end{pgfonlayer}
\end{tikzpicture}}
\endpgfgraphicnamed} 
\end{center}
\caption{An example of updates by functions $F,G,I$  on an entity-relation cube}
\label{fig:egcube}
\end{figure}
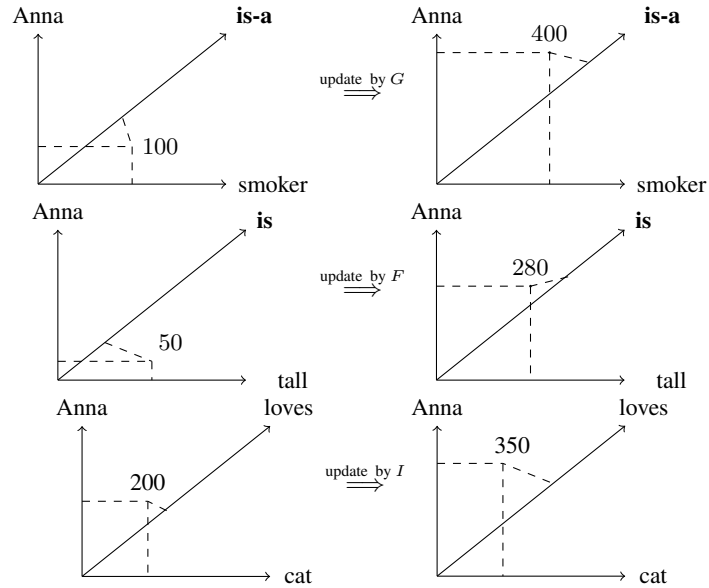

As an example, consider the series of updates depicted in Figure \ref{fig:egcube}. 

Relative pronouns such as `who' update the entry corresponding to the head of the relative clause and the rest of the clause.  For example in the clause  `the man who went home', we update $c_{ijk}$ for $i$ the index of `man' as subject of the verb `went' with index $j$ and its object `home' with index $k$.  Propositional attitudes such as `know' update the entry value of $c_{ijk}$ for $i$ the index of their subject, $j$ the index of themselves, and $j$ the index of their proposition. For instance in the sentence `John knows Mary slept', we update the entry value for `John', `know' and the proposition `Mary slept'.  The conjunctive \emph{and} is  modelled as before.  

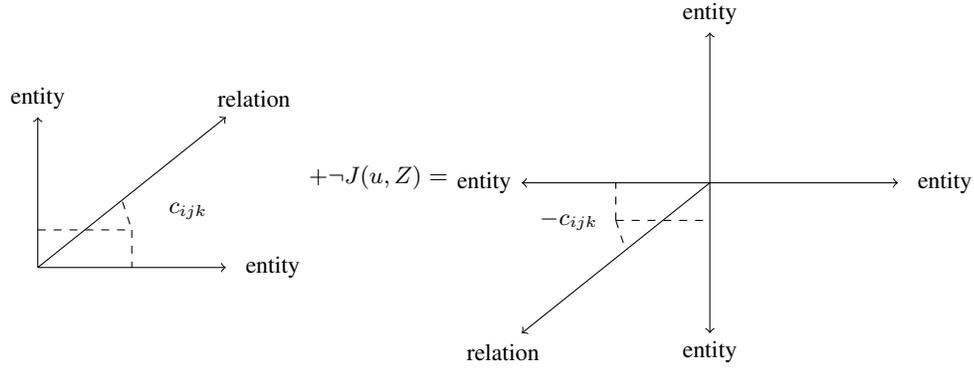
\begin{figure}[t!]

{%
\beginpgfgraphicnamed{cube1}
\begin{tikzpicture}
	\begin{pgfonlayer}{nodelayer}
		\node [style=none] (0) at (-5, 3) {};
		\node [style=none] (1) at (-5, -1) {};
		\node [style=none] (2) at (0, -1) {};
		\node [style=none] (3) at (0, 3) {};
		\node [style=none] (4) at (-2.5, 0) {};
		\node [style=none] (5) at (-2.5, -1) {};
		\node [style=none] (6) at (-5, 0) {};
		\node [style=none] (7) at (-2.75, 0.75) {};
		\node [style=none] (8) at (-1, 0.5) {$c_{ijk}$};
		\node [style=none] (9) at (-5, 3.5) {entity};
		\node [style=none] (10) at (0.75, 3.5) {relation};
		\node [style=none] (11) at (1.25, -1) {entity};
	\end{pgfonlayer}
	\begin{pgfonlayer}{edgelayer}
		\draw [<-] (0.center) to (1.center);
		\draw [->] (1.center) to (2.center);
		\draw [->] (1.center) to (3.center);
		\draw [dashed] (6.center) to (4.center);
		\draw [dashed] (4.center) to (5.center);
		\draw [dashed] (7.center) to (4.center);
	\end{pgfonlayer}
\end{tikzpicture}}
\endpgfgraphicnamed} 
$  +  \neg J(u,Z)  =  $
{%
\beginpgfgraphicnamed{cube-neg-minus}
\begin{tikzpicture}
	\begin{pgfonlayer}{nodelayer}
		\node [style=none] (0) at (0, -4) {};
		\node [style=none] (1) at (0, 0) {};
		\node [style=none] (2) at (-5, 0) {};
		\node [style=none] (3) at (0, -4.5) {entity};
		\node [style=none] (4) at (-6, 0) {entity};
		\node [style=none] (5) at (-5, -4) {};
		\node [style=none] (6) at (-3.75, -1) {$- c_{ijk}$};
		\node [style=none] (7) at (-5.5, -4.5) {relation};
		\node [style=none] (8) at (0, -1) {};
		\node [style=none] (9) at (-2.25, -1.75) {};
		\node [style=none] (10) at (-2.5, -1) {};
		\node [style=none] (11) at (-2.5, 0) {};
		\node [style=none] (12) at (5, 0) {};
		\node [style=none] (13) at (0, 4) {};
		\node [style=none] (14) at (0, 4.5) {entity};
		\node [style=none] (15) at (6.25, 0) {entity};
	\end{pgfonlayer}
	\begin{pgfonlayer}{edgelayer}
		\draw [<-] (0.center) to (1.center);
		\draw [->] (1.center) to (2.center);
		\draw [->] (1.center) to (5.center);
		\draw [dashed] (10.center) to (8.center);
		\draw [dashed] (10.center) to (11.center);
		\draw [dashed] (10.center) to (9.center);
		\draw [->] (1.center) to (12.center);
		\draw [->] (1.center) to (13.center);
	\end{pgfonlayer}
\end{tikzpicture}}
\endpgfgraphicnamed}

%
\caption{The cube of an example negation operator}
\label{fig:negation}
\end{figure}

Negation can be modelled  by  assigning the abstract type $D(DS)S$ to it and a term homomorphism $\lambda uZ.
\neg J( u, Z, c)$ with the object type $(V(VU))U$.  Concretely, it   negates  the value entry of its input entity relation triple by, for example,  setting  $\neg J (u,Z, c)  :=  - J (u,Z, c)$, as shown in Figure \ref{fig:negation}.

\begin{figure}[h]
\begin{center}
{%
\beginpgfgraphicnamed{cube1Anna}
\begin{tikzpicture}
	\begin{pgfonlayer}{nodelayer}
		\node [style=none] (0) at (-5, 3) {};
		\node [style=none] (1) at (-5, -1) {};
		\node [style=none] (2) at (0, -1) {};
		\node [style=none] (3) at (0, 3) {};
		\node [circle, fill, style=none] (4) at (-2, 2) {};
		\node [style=none] (5) at (-2, -1) {};
		\node [style=none] (6) at (-5, 2) {};
		\node [style=none] (7) at (-1, 2.25) {};
		\node [style=none] (8) at (-2, 2.5) {$380$};
		\node [style=none] (9) at (-5, 3.5) {Anna};
		\node [style=none] (10) at (1, 3.5) {\bf is-a};
		\node [style=none] (11) at (1.25, -1) {woman};
	\end{pgfonlayer}
	\begin{pgfonlayer}{edgelayer}
		\draw [<-] (0.center) to (1.center);
		\draw [->] (1.center) to (2.center);
		\draw [->] (1.center) to (3.center);
		\draw [dashed] (6.center) to (4.center);
		\draw [dashed] (4.center) to (5.center);
		\draw [dashed] (7.center) to (4.center);
	\end{pgfonlayer}
\end{tikzpicture}}
\endpgfgraphicnamed} $\stackrel{\text{update \ by\ } G}{\implies}${%
\beginpgfgraphicnamed{cube2AnnaU}
\begin{tikzpicture}
	\begin{pgfonlayer}{nodelayer}
		\node [style=none] (0) at (-5, 2.5) {Anna};
		\node [style=none] (1) at (-3, 2) {351};
		\node [style=none] (2) at (0.5, 2.5) {loves};
		\node [style=none] (3) at (-3, 1.25) {};
		\node [style=none] (4) at (-5, -2) {};
		\node [style=none] (5) at (-5, 1.25) {};
		\node [style=none] (6) at (-5, 2) {};
		\node [style=none] (7) at (0.75, -2) {cat};
		\node [style=none] (8) at (0, -2) {};
		\node [style=none] (9) at (0, 2) {};
		\node [style=none] (10) at (-3, -2) {};
		\node [style=none] (11) at (-2, 0.5) {};
	\end{pgfonlayer}
	\begin{pgfonlayer}{edgelayer}
		\draw [<-] (6.center) to (4.center);
		\draw [->] (4.center) to (8.center);
		\draw [->] (4.center) to (9.center);
		\draw [dashed] (5.center) to (3.center);
		\draw [dashed] (3.center) to (10.center);
		\draw [dashed] (11.center) to (3.center);
	\end{pgfonlayer}
\end{tikzpicture}}
\endpgfgraphicnamed} 

\smallskip
{%
\beginpgfgraphicnamed{cube1Susan}
\begin{tikzpicture}
	\begin{pgfonlayer}{nodelayer}
		\node [style=none] (0) at (5.5, 3) {};
		\node [circle, fill, style=none] (1) at (3.5, 2) {};
		\node [style=none] (2) at (4.5, 2.25) {};
		\node [style=none] (3) at (0.5, 3) {};
		\node [style=none] (4) at (6.5, 3.5) {\bf is-a};
		\node [style=none] (5) at (3.5, -1) {};
		\node [style=none] (6) at (0.5, 3.5) {Susan};
		\node [style=none] (7) at (0.5, -1) {};
		\node [style=none] (8) at (3.5, 2.5) {$380$};
		\node [style=none] (9) at (6.75, -1) {woman};
		\node [style=none] (10) at (5.5, -1) {};
		\node [style=none] (11) at (0.5, 2) {};
	\end{pgfonlayer}
	\begin{pgfonlayer}{edgelayer}
		\draw [<-] (3.center) to (7.center);
		\draw [->] (7.center) to (10.center);
		\draw [->] (7.center) to (0.center);
		\draw [dashed] (11.center) to (1.center);
		\draw [dashed] (1.center) to (5.center);
		\draw [dashed] (2.center) to (1.center);
	\end{pgfonlayer}
\end{tikzpicture}}
\endpgfgraphicnamed} $\stackrel{\text{update \ by\ } F}{\implies}${%
\beginpgfgraphicnamed{cube2SusanU}
\begin{tikzpicture}
	\begin{pgfonlayer}{nodelayer}
		\node [style=none] (0) at (0.5, -2) {};
		\node [style=none] (1) at (0.5, 2.5) {Susan};
		\node [style=none] (2) at (2.75, 1.5) {320};
		\node [style=none] (3) at (6.25, -2) {cat};
		\node [style=none] (4) at (2.5, -2) {};
		\node [style=none] (5) at (6, 2.5) {loves};
		\node [style=none] (6) at (3.25, 0.25) {};
		\node [style=none] (7) at (5.5, 2) {};
		\node [style=none] (8) at (0.5, 2) {};
		\node [style=none] (9) at (2.5, 1) {};
		\node [style=none] (10) at (0.5, 1) {};
		\node [style=none] (11) at (5.5, -2) {};
	\end{pgfonlayer}
	\begin{pgfonlayer}{edgelayer}
		\draw [<-] (8.center) to (0.center);
		\draw [->] (0.center) to (11.center);
		\draw [->] (0.center) to (7.center);
		\draw [dashed] (10.center) to (9.center);
		\draw [dashed] (9.center) to (4.center);
		\draw [dashed] (6.center) to (9.center);
	\end{pgfonlayer}
\end{tikzpicture}}
\endpgfgraphicnamed} 

\smallskip
{%
\beginpgfgraphicnamed{cube1Aunt}
\begin{tikzpicture}
	\begin{pgfonlayer}{nodelayer}
		\node [style=none] (0) at (5.5, 3) {};
		\node [circle, fill, style=none] (1) at (4.25, 2.75) {};
		\node [style=none] (2) at (4.5, 2.25) {};
		\node [style=none] (3) at (0.5, 3) {};
		\node [style=none] (4) at (6.5, 3.5) {\bf is-a};
		\node [style=none] (5) at (4.25, -1) {};
		\node [style=none] (6) at (0.5, 3.5) {Aunt};
		\node [style=none] (7) at (0.5, -1) {};
		\node [style=none] (8) at (4.25, 3.25) {$500$};
		\node [style=none] (9) at (6.75, -1) {woman};
		\node [style=none] (10) at (5.5, -1) {};
		\node [style=none] (11) at (0.5, 2.75) {};
	\end{pgfonlayer}
	\begin{pgfonlayer}{edgelayer}
		\draw [<-] (3.center) to (7.center);
		\draw [->] (7.center) to (10.center);
		\draw [->] (7.center) to (0.center);
		\draw [dashed] (11.center) to (1.center);
		\draw [dashed] (1.center) to (5.center);
		\draw [dashed] (2.center) to (1.center);
	\end{pgfonlayer}
\end{tikzpicture}}
\endpgfgraphicnamed} $\stackrel{\text{update \ by\ } I}{\implies}${%
\beginpgfgraphicnamed{cube2AuntU}
\begin{tikzpicture}
	\begin{pgfonlayer}{nodelayer}
		\node [style=none] (0) at (2.75, 1.5) {};
		\node [style=none] (1) at (0.5, -2) {};
		\node [style=none] (2) at (0.5, 2) {};
		\node [style=none] (3) at (0.5, 1.5) {};
		\node [style=none] (4) at (5.5, -2) {};
		\node [style=none] (5) at (6, 2.5) {loves};
		\node [style=none] (6) at (2.75, -2) {};
		\node [style=none] (7) at (2.75, 2) {410};
		\node [style=none] (8) at (5.5, 2) {};
		\node [style=none] (9) at (4, 0.75) {};
		\node [style=none] (10) at (6.25, -2) {cat};
		\node [style=none] (11) at (0.5, 2.5) {Aunt};
	\end{pgfonlayer}
	\begin{pgfonlayer}{edgelayer}
		\draw [<-] (2.center) to (1.center);
		\draw [->] (1.center) to (4.center);
		\draw [->] (1.center) to (8.center);
		\draw [dashed] (3.center) to (0.center);
		\draw [dashed] (0.center) to (6.center);
		\draw [dashed] (9.center) to (0.center);
	\end{pgfonlayer}
\end{tikzpicture}}
\endpgfgraphicnamed} 
\end{center}
\caption{An example of the updates incurred by  the universal quantifier}
\label{fig:egcubeall}
\end{figure}
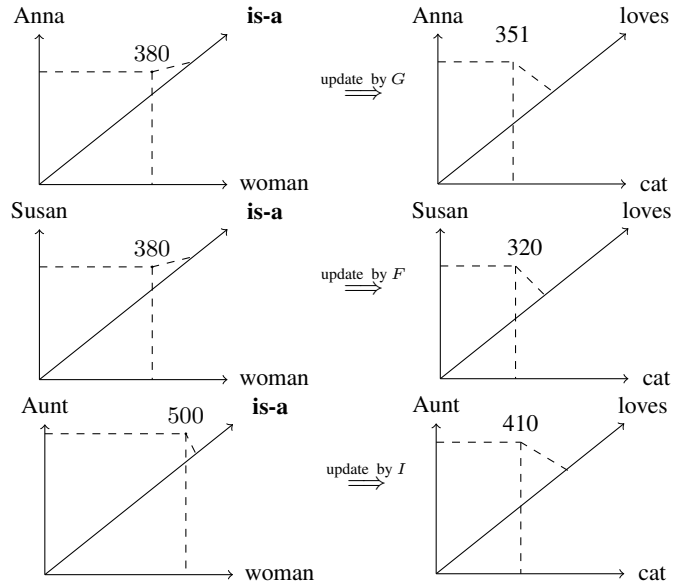

Quantifiers can be modelled in a more elaborated way. For instance  \emph{every} can have the lambda term $\lambda uZ. \forall(u, Z, c)$  assigned to it, which finds {all} the  entries of the  graph that have an {\sf is-a} relation with $u$,  then updates their value entries where they relate to the verb phrase $Z$  by increasing them. An example is depicted in Figure \ref{fig:egcubeall}, regarding the universal quantifier  in a  sentence such as ``All women love cats''. Here, it finds all entities that are in an {\sf is-a} relation with \emph{woman}, such as ``Anna {\sf is-a} woman'', ``Susan {\sf is-a} woman'', ``Aunt {\sf is-a} woman''. Then it increases $c_{ijk}$ where $i$ is the index of ``Anna'', $j$ is the index of ``love'' and $k$ is the index of ``cat'', and similarly for ``Susan'' and ``Aunt''.   The quantifier  \emph{some}, acts in a similar way with the difference that it  only updates {some} of these value entries, e.g. by random choice. For example, in the above case, it only increases the entry  corresponding to `Anna' and leaves `Aunt''  and ``Susan'' unchanged.   Generalised  quantifiers act in  similar ways, e.g. \emph{most} changes {most} of its corresponding entries, \emph{at least 5} changes {at least 5} of them; \emph{at most 5} changes {at most 5} of them (all chosen randomly)  and so on.

%
%
%
%


These update instructions   fall through the semantics of phrases and
sentences compositionally as in the case of co-occurrence matrices.

\section{A Logic for Context Change Potentials}
A logic for sentences as context change potentials has a syntax  as follows: 

\[
\phi ::= p \mid \neg \phi \mid \phi \wedge \psi \mid 
\]
with disjunction and implication operations  defined using de Morgan duality:

\begin{eqnarray*}
\phi \vee \psi &:=& \neg (\neg \phi \wedge \neg \psi)\\
\phi \to \psi &:=& \neg \phi \vee \psi
\end{eqnarray*}

\noindent
This logic is the propositional fragment of the logic  of context change potentials, presented   in \cite{MuskensBenthemVisser}, based on the ideas of Heim \cite{Heim1983}. Heim   extends the theory of presupossitions of Karttunen \cite{Karttunen1974} and defines the context change potential of a  sentence as a function of the context change potentials of its parts, an idea that leads to the development of the above logic. The logic we consider here is the same logic but without the presupposition operation.  

We refer to the language of this logic as ${\cal L}_{ccp}$.    For a context $c$, a context change potential is defined as follows:

\begin{eqnarray*}
||p||(c) &:=& c + || p ||
\\
||\neg \phi ||(c) &:=& c - ||\phi||(c)\\
||\phi \wedge \psi || &:=& ||\psi||(||\phi||(c))
\end{eqnarray*}

It is easy to verify that:

\begin{eqnarray*}
|| \phi \vee \psi || &=& ||\psi||(c) - ||\psi||(||\phi||(c))\\
|| \phi \to \psi||(c) &=& c - (||\phi ||(c) - ||\psi ||(||\phi||(c)))
\end{eqnarray*}

Here,  $||\phi ||$ is the context change potential of $\phi$ and  a function from contexts to contexts. 
Whereas for Heim,  contexts and context change potentials of atomic sentences $||p||$ are both sets of valuations, for us contexts are co-occurrence matrices or entity relation cubes and context change potentials of atomic sentences are vectors. Thus,  where the context change potential operation of Heim simply takes the intersection of a context and a context change potential  $c \cap ||p ||$, we have to do an operation that acts on  matrices/cubes  rather than sets. We use the update operation   of  term homomorphisms, defined in the previous sections,   and define a context change potential as follows:

\begin{definition}
For $S$ a sentence  in  ${\cal L}_{ccp}$,  $||S||$ its context change potential,  $H(S)$ the term homomorphic image of $S$, and 
 $c$ a co-occurrence matrix  or an  entity relation  cube, we  define:
\begin{eqnarray*}
||S||(c) &:=& c +'  H(S)\\
c -  H(S) &:=& (c +' H(S))^{-1} 
\end{eqnarray*}
 for $+'$  the   update operation  defined on term homomorphisms  and $-'$ its inverse, defined as follows for matrices:
 

{\small
\begin{eqnarray*}
\left ( \begin{array}{ccc}
m_{11} & \cdots & m_{1k}\\
m_{21} &  \cdots & m_{2k}\\
\vdots&&\\
m_{n1} &  \cdots & m_{nk}
\end{array} \right) 
 &+'&  \langle \ov{a}, u,v, \cdots \rangle = 
\left ( \begin{array}{ccc}
m'_{11} & \cdots & m'_{1k}\\
m'_{21} &  \cdots & m'_{2k}\\
\vdots&&\\
m'_{n1} &  \cdots & m'_{nk}
\end{array} \right)  \quad  \text{for} \quad m'_{ij} := \begin{cases} 1 & m_{ij}=1 \\ 1 & m_{ij} = 0\end{cases}
\\ &&\\
\left ( \begin{array}{ccc}
m_{11} & \cdots & m_{1k}\\
m_{21} &  \cdots & m_{2k}\\
\vdots&&\\
m_{n1} &  \cdots & m_{nk}
\end{array} \right) 
 &-'&  \langle \ov{a}, u,v, \cdots \rangle = 
\left ( \begin{array}{ccc}
m'_{11} & \cdots & m'_{1k}\\
m'_{21} &  \cdots & m'_{2k}\\
\vdots&&\\
m'_{n1} &  \cdots & m'_{nk}
\end{array} \right) \quad \text{for} \quad m'_{ij} := \begin{cases} 0 & m_{ij}=1 \\ 0 & m_{ij} = 0\end{cases}
\end{eqnarray*}
}
The definition of $+'$ and $-'$ for cubes are done similarly. 
\end{definition}

 The $+'$ operation updates the co-occurrence matrix in a binary fashion: if the entry $m_{ij}$ of the matrix has already been updated, and thus has value 1, then a succeeding update is not going to increase the value from 1 to 2 and  will keep it as 1. Conversely, when the $-'$ operation acts on an entry $m_{ij}$ which is already 0, it is not going to change its value, but if it acts on a non-zero $m_{ij}$, that is an $m_{ij}$ which has value 1, it will decrease it to 0. The procedure is similar  for the cubes. The resulting matrices and cubes will have binary entires, that is, they will either be 1's or 0's. A 1 indicates that at least one occurrence of the roles associated to the entires have been seen in the corpus before; a  0 indicates that none has been seen or that a role and its negation  have occurred. 
 
 Fixing a bijection between the elements $[1,n] \times [1,k]$ of our matrices and natural numbers $[1,n\times k]$ and between elements $[1,n] \times [1,k] \times [1,z]$ of the cubes and natural numbers $[1,n\times k \times z]$, one can show that   $c +' H(S)$ is the table of a  binary relation in the case of matrices and ternary relation in the case of cubes. Those entries $(i,j)$ of the matrices and $(i,j,k)$ of the cubes  that have a non zero value entry, are mapped to an element of the relation. An example of this isomorphism is as shown below for a 2 $\times $ 2 matrix:
 \[
 \left (\begin{array}{cc} 1 & 0 \\ 1 & 1 \end{array} \right ) 
  \mapsto
   \quad \mapsto  \quad 
 \begin{tabular}{c|cc}
 & 1 & 2\\ \hline 
 1 & 1& 0\\ 
 2& 1&1\end{tabular} \quad
 \quad \{(1,1), (2,1), (2,2)\}
 \] 
  
 These  binary updates  can be seen as providing  a notion of   `contextual truth', that is,  for example,  a sentence $S$  is true in a given a context  $c$, whenever the update resulting from $s$ is  already included in the matrix or cube of its context, i.e. its update  is one that does not change $c$.

As argued  in  \cite{MuskensBenthemVisser}, the semantics of this logic is dynamic, in the sense that the context change potential  of a sequence of sentences is obtained by function composition, that is as follows:
\[
||S_1, \cdots, S_n  ||(c) := ||S_1|| \circ \cdots \circ ||S_n||(c)
\]

Using this dynamic semantics, it is straightforward to  show that 
\begin{proposition}
The context  $c$ corresponding to  the sequence of sentences  $S_1, \cdots, S_n$,  is  the zero vector  updated by that sequence of sentences, that is:
\[
c = ||S_1, \cdots, S_n  ||(\ov{0})
\]
where 
\begin{eqnarray*}
||S||(c) &:=& c + H(S)\\
c -  H(S) &:=& (c + H(S))^{-1} 
\end{eqnarray*}
\end{proposition}

In the case of the  co-occurrence matrices, $c$ is the co-occurrence matrix and $\ov{0}$ is the zero matrix. In the entity relation cube case, $c$ is the entity relation  cube and $\ov{0}$ is the zero cube. We are  using the usual real number addition  and subtraction on $m_{ij}$ and $c_{ijk}$ entires of the matrices and cubes, that is 
\begin{align*}
m'_{ij} := m_{ij} + 1 \qquad &
m'_{ij} := m_{ij} -1\\
c'_{ijk} := c_{ijk} + 1 \qquad &
c'_{ijk} := c_{ijk} -1
\end{align*}

We will refer to a sequence of sentences as a \emph{corpus}.

\section{Admittance of  Sentences by Contexts}

A notion of \emph{admittance of a sentence by a context} was developed by Karttunen for presuppositions and extended by Heim for context change potentials. It is defined as follows, for $c$ a context and $\phi$ a proposition of  ${\cal L}_{ccp}$: 
\[
\mbox{context} \ c \ \mbox{admits proposition}  \  \phi  \ \iff \  ||\phi ||(c) = c
\]

We  use  this notion   and  develop a similar notion between  a corpus and a sentence. 

\begin{definition}
A corpus admits a sentence  iff    the context $c$ (a co-occurrence matrix or entity relation  cube) built from it, admits  it. 
\end{definition}

Consider the  following corpus:

`Cats and dogs are animals that sleep. Cats chase cats and mice. Dogs chase all animals. Cats like mice, but mice fear cats, since cats eat mice. Cats smell mice and mice run   from cats.'

It admits the following sentences:

\begin{quote}
Cats are animals.\\
Dogs are animals.\\
Cats chase cats.\\
Cats chase mice.\\
Dogs chase cats and dogs.
\end{quote}

\noindent
Note that this notion of admittance caters for monotonicity of inference. For instance, in the above example, from the sentences ``Cats [and dogs] are animals [that sleep]" and ``Dogs chase all animals", we can infer that the context admits the sentence "Dogs chase cats". 

On the other hand, $c$ does not admit the negation of the above, for example  it does not admit

\begin{quote}
(*) Dogs do not chase cats.\\
(*) Dogs do not chase  dogs.
\end{quote}

It also do not admit the negations of derivations of the above or negations of sentences of the cropus, for example, it does not admit

\begin{quote}
(*) Cats are not animals.\\
(*) Dogs do not sleep.
\end{quote}

The corpus misses a sentence asserting that mice are also animals. Hence, $c$  does not admit the sentence `dogs chase mice'. Some other sentences that are not admitted by $c$ as as follows:

\begin{quote}
(*) Cats like dogs.\\
(*) Cats eat dogs.\\
(*) Dogs run from cats.\\
(*) Dogs like mice.\\
(*) Mice fear dogs.\\
(*) Dogs eat mice.
\end{quote}

One can argue that by binarizing the update operation and using $+'$ and $-'$ rather than the original $+$ and $-$, we are loosing the full power of distributional semantics.  It seems wasteful to  rather than building context matrices by counting co-occurrences, only record if something co-occurred with another or not. This can be overcome by working with a pair of contexts: a binarized one and a numerical one.  The binarized context allows for defining a notion of admittance as before, and the numerical one allows to use numerical values, e.g. the degrees of similarity between words.  The notion of word similarity used in distributional semantics  is a direct consequence of the distributional hypothesis:  it says that  words that often occur in the same contexts have similar meanings \cite{Firth1957}. Various formal notions have been used to measure the above degree of similarity, amongst the  successful ones is the cosine of the angle between the vectors of the words.  If the vectors are normalised to have length 1, which we shall assume, cosine becomes the same as the dot product of the vectors.  Then, one can use these degrees of similarity to assign a numerical value to the admittance relation, e.g. as follows:

\begin{quote}
A pair of binary and numerical co-occurrence matrices $c$ and $c'$admit a sentence $s'$ with degree $d$, if $c$ admits  $s$,  and $s'$ is obtained from $s$ by replacing a word $w$ of $s$ with a word  $w'$ such that $w'$ has the same grammatical role in $s'$ as $w$  in $s$ and the  degree of similarity between  $w$ and $w'$ is $d$, computed from the numerical entries of $c'$.  
\end{quote}
Here,  $c$ admits $s$ and if there is a word in $s$  that is similar to another word $w'$, then if we replace $w$ in $s$ with $w'$ (keeping the grammatical role that $w$ had in $s$) then the sentence resulting from this substitution, i.e. $s'$ is also admitted by $c$, albeit with a degree equal to the degree of similarity between $w$ and $w'$.  This degree is computed using the numerical values recored in $c'$. The above can be extended to the case when one  replaces more than one word in $s$ with words similar to them. Then,  the degree of entailment  may be obtained by multiplying the degrees of similarities of the individually replaced words. 

%

The normalised context matrix of our example corpus above are as  in tables \ref{table:example-matrix}, where for simplicity the  co-occurrence window is taken to be ``occurrence within the same sentence''.  


\begin{table}[t!]
\begin{center}
\begin{tabular}{c|cccccccc}
& 1&2&3&4&5&6 & 7 & 8 \\
&animal&sleep&  chase & like & fear & eat &smell & run\\
\hline
&\\
1- cats& $\frac{1}{8}$ &$\frac{1}{8}$  & $\frac{1}{8}$&$\frac{1}{8}$ & $\frac{1}{8}$& $\frac{1}{8}$ & $\frac{1}{8}$ & $\frac{1}{8}$  \\
&\\
2- mice& 0 & 0 & $\frac{1}{6}$ & $\frac{1}{6}$&$\frac{1}{6}$ & $\frac{1}{6}$ & $\frac{1}{6}$ & $\frac{1}{6}$  \\
&\\
3- dogs &$\frac{1}{2}$ & $\frac{1}{4}$ & $\frac{1}{4}$ & 0 & 0 &0 & 0 & 0 \\
\end{tabular}
\end{center}
\caption{The normalised co-occurrence matrix built from the example corpus with the co-occurrence window taken to be occurrence within the same sentence.}
\label{table:example-matrix}
\end{table}


From the context matrix one obtains the following degrees of similarity:
\begin{eqnarray*}
\cos(\text{cats}, \text{mice}) &=& 6 \times (\frac{1}{6} \times \frac{1}{8}) = \frac{1}{8}\\
\cos(\text{cats}, \text{dogs}) &=& (\frac{1}{2}) \times 2 \times (\frac{1}{4} \times \frac{1}{8})  = \frac{1}{32}\\
\cos(\text{dogs}, \text{mice})  &=& \frac{1}{4} \times \frac{1}{6} = \frac{1}{24}
\end{eqnarray*}

The corpus misses an explicit sentence declaring that mice are also animals. Hence, from the sentences of the  corpus the negation of `dogs chase mice' follows, which is a wrong entailment in the real world. This wrong can now be put right, since we can  replace  the word `Cats' in the admitted sentence `Cats are animals'  with `Mice'; as we have $\cos(\text{cats}, \text{mice}) = \frac{1}{8}$, thus obtaining  the situation where $c$ admits  the following, both with degree $\frac{1}{8}$:

\begin{quote}
Mice are animals.\\
Dogs chase mice.
\end{quote} 
These were  not possible before. 
We also obtain admittance of  the following  sentences albeit with a lower degree of $\frac{1}{24}$:
\begin{quote}
(*) Cats like dogs.\\
(*) Cats eat dogs.\\
(*) Dogs run from cats.
\end{quote}
Some other examples are as follows with a still lower degree of $\frac{1}{32}$:

\begin{quote}
(*) Dogs like mice.\\
(*) Mice fear dogs.\\
(*) Dogs eat mice.
\end{quote}

 Some of the above are as likely as the ones that were derived with degree $\frac{1}{8}$. This is since the degrees come from co-occurrences in corpora and the one that we have is quite restrictive. One hopes that the bigger a corpus,  the more reflective of the real world it is. Another way of improving the word-based entailments is by using     linguistic resources such as WordNet, e.g. replacing words with their hypernyms. 

\subsection{Evaluating  on Existing Entailment Datasets}
It remains to show if the notion of admittance of a sentence by a context  can be applied to derive  entailment relations between sentences.  In future work, we will put this method to test on the main and down stream   inference datasets such as FraCas \cite{Cooper1996}, SNLI\cite{SNLI2015},  Zeichner\cite{Zeichner2012} and datasets of the RTE challenge.  The FraCas inferences are logical and the lambda calculus models of language should help in   deriving them. As an example,  consider the  \texttt{fracas-013} test case: 
\begin{quote}
\texttt{fracas-013} \quad 	answer: yes\\
$P1$\quad Both leading tenors are excellent.\\
$P2$	\quad Leading tenors who are excellent are indispensable.\\
$Q$\quad	\ \ Are both leading tenors indispensable?\\
$H$\quad	\ \  Both leading tenors are indispensable.\\
\end{quote}
In our setting, using the updates resulting from P1 and P2, one can contextually derive H. 
Zeichner however does take the similarity between words into account. An example is the following entailment between  two sentences; this entailment  was judged to be true with confidence by human annotators:

\begin{quote}
Parents have a great influence on the career development of their children.\\
Parents have a powerful influence on the career development of their children.
\end{quote} 

We can derive the above with a contextual entailment consisting of a cube updated by just the above two sentences, with  the degree of similarity between `powerful' and `great', mined from the co-occurrence matrix of a large corpus. 

The judgements  of the SNLI dataset are more tricky as they rely on external knowledge. For example consider the  entailment between the following sentences:

\begin{quote}
A soccer game with multiple males playing.\\
Some men are playing a sport.
\end{quote}
or the contradiction between the following:

\begin{quote}
A black race car starts up in front of a crowd of people.\\
A man is driving down a lonely road.
\end{quote}

Deciding these correctly is a challenge for our framework. The strength of our approach is in deciding whether a set of  sentences follow from a given corpus of text, rather than in judging entailment relations between a given  pair or triple of sentences.  We shall, nevertheless,  try to experiments with all these datasets. 

\section{Conclusion and Future Directions}
We showed how a static interpretation of
a lambda calculus model of natural language  provides vector representations  for phrases and sentences. Here, the  type of the vector of a word depended on
its abstract type and could be an atomic vector, a matrix, or a cube,
or a tensor of higher rank. Combinations of these  vary
based on the tensor rank of the type of each word involved in the combination. 
For instance,  one
could take the matrix multiplication of the matrix of an intransitive
verb with the vector of its subject, whereas for a transitive verb the
sequence of operations were a contraction between the cube of the verb
and the vector of its object followed by a matrix multiplication
between the resulting matrix and the vector of the subject.  A toolkit
of functions needed to perform these operations was defined. This toolkit can be restated  for the higher order types, such  $I^2R$ and $I^3R$,
rather than the current $IR$, to provide means of combining matrices, cubes, 
and their updates, if needed. 

We extended the above setting by  reasoning about the notion of context and its update and developing  a dynamic vector  interpretation for the language of  lambda terms. Truth
conditional and vector models of language follow two very different
philosophies. The vector models are based on contexts, the truth
models on denotations. Our first interpretation was static and based on truth conditions. Our second approach is  based on a dynamic interpretation, where we followed the context update model of Heim, and  hence, is deemed 
the more appropriate choice. We showed how Heim's
files can be turned into vector contexts and how her context change
potentials can be used to provide vector interpretations for phrases
and sentences.  We  treated sentences as Heim's `context change potentials' and provided update instructions  for words therein--including  quantifiers, negation, and coordination words. We provided two concrete realisations of  contexts: co-occurrence matrices and entity relation cubes and in each case detailed how  these  context update instructions allow contexts thread through vector semantics in a compositional manner. With an eye towards a large scale empirical evaluation of the model, we defined a notion of   `contexts admitting sentences' and  degrees thereof between contexts and sentences and showed, by means of examples, how these can be used to judge whether a  sentence is entailed by a cube context or a pair of cube and matrix contexts.  A large scale empirical evaluation of the model constitutes work in progress.


%
Our approach is applicable to the lambda terms obtained via other
syntactic models, e.g. CCG, and Lambek Grammars and can also be modified  to develop a vector semantics for LFG. We also aim to work with other update
semantics, such as continuation-based approaches.  One could also have
a general formalisation wherein both the static approach of previous
work and the dynamic one of this work cohabit. This can be done by
working out a second pair of type-term homomorphisms that will also
work with Heim's possible world part of the contexts. In this setting,
the two concepts of meaning: truth theoretic and contextual, each with
its own uses and possibilities, can work in tandem. 

 An intuitive  connection to Fuzzy logic is imaginable,  wherein one  interprets the logical words in more sophisticated ways, for instance,  conjunction and disjunction   take max and  min of their entires, or add and  subtract them. It may be worth investigating  if  such connections  add to the applicability of the current model  and if so making the connection formal. 

\subsubsection*{Acknowledgements} We wish to thank the anonymous
referees of a short  version of this paper presented in LACL 2017  for excellent feedback. The research done for this paper was supported by the Royal Society International Exchange Award IE161631.

\bibliographystyle{splncs03}
\bibliography{ref}
\end{document}